\documentclass[letterpaper, 10 pt, conference]{ieeeconf}  

\IEEEoverridecommandlockouts                              

\usepackage{graphicx}
\usepackage{tikz}
\usepackage{xcolor}
\usepackage{graphicx}
\usepackage{etoolbox}
\usepackage{booktabs}
\usepackage{multirow}
\usepackage{amsmath}
\usepackage{verbatim}
\usepackage{amsfonts}
\usepackage{hyperref}

\definecolor{wacvblue}{rgb}{0.21,0.49,0.74}
\definecolor{clustvitorange}{rgb}{1,0.439,0.263}
\definecolor{clustvitgreen}{rgb}{0,0.537,0.482}
\DeclareRobustCommand{\numberdotorange}[1]{%
  \tikz[baseline=(char.base)]{
    \node[shape=circle, fill=clustvitorange, text=white, inner sep=1pt, font=\small\bfseries] (char) {#1};
  }%
}

\DeclareRobustCommand{\numberdotgreen}[1]{%
  \tikz[baseline=(char.base)]{
    \node[shape=circle, fill=clustvitgreen, text=white, inner sep=1pt, font=\small\bfseries] (char) {#1};
  }%
}

\title{\LARGE \bf
ClustViT: Clustering-based Token Merging for Semantic Segmentation
}

\author{Fabio Montello, Ronja Güldenring and Lazaros Nalpantidis 
\thanks{This work has been supported by Innovation Fund Denmark through the project ``Safety and Autonomy for Vehicles in Agriculture (SAVA)", 2105-00013A.}
\thanks{All authors are with the Department of Electrical and Photonics Engineering, DTU - Technical University of Denmark, Kgs. Lyngby, Denmark.}
\thanks{{\tt\small \{fabmo, ronjag, lanalpa\}@dtu.dk}}}

\begin{document}

\maketitle
\thispagestyle{empty}
\pagestyle{empty}


\begin{abstract}
Vision Transformers can achieve high accuracy and strong generalization across various contexts, but their practical applicability on real-world robotic systems is limited due to their quadratic attention complexity. Recent works have focused on dynamically merging tokens according to the image complexity. Token merging works well for classification but is less suited to dense prediction. We propose ClustViT, where we expand upon the Vision Transformer (ViT) backbone and address semantic segmentation. Within our architecture, a trainable Cluster module merges similar tokens along the network guided by pseudo-clusters from segmentation masks. Subsequently, a Regenerator module restores fine details for downstream heads. Our approach achieves up to $2.18\times$ fewer GFLOPs and $1.64\times$ faster inference on three different datasets, with comparable segmentation accuracy. Our code and models are made publicly available\footnote{\href{https://github.com/DTU-PAS/clustvit}{https://github.com/DTU-PAS/clustvit}}.


 \end{abstract}

\section{Introduction}

\label{sec:intro}
Reducing the computational cost of Vision Transformers (VIT) \cite{dosovitskiyImageWorth16x162021b} in dense prediction tasks, such as semantic segmentation, is essential when considering autonomous robotic systems that need to perceive their operational environments.
The Transformer architecture \cite{vaswaniAttentionAllYou2017} is the de facto architecture of choice when it comes to computer vision solutions that require high performance across different contexts \cite{carionEndtoEndObjectDetection2020, zhangSegViTSemanticSegmentation}. However, one major limitation of Transformers is that their computational complexity grows quadratically with respect to the input size.
Even though existing approaches have successfully reduced token redundancy for classification tasks, they often struggle to generalize to dense prediction settings, where preserving spatial and semantic detail is critical.

In this paper, we focus on semantic segmentation and argue that the computational cost of ViTs can be significantly reduced without compromising segmentation accuracy by incorporating appropriate token clustering. To achieve this, we propose making dual use of ground truth segmentation masks. More precisely, the semantic information embedded in segmentation masks is not only used to supervise the segmentation task. We also use it to train a token clustering component integrated between Transformer blocks of a ViT backbone, which allows the model to identify and merge semantically similar tokens.

\begin{figure}[t]
  \centering
  \includegraphics[width=.9\linewidth]{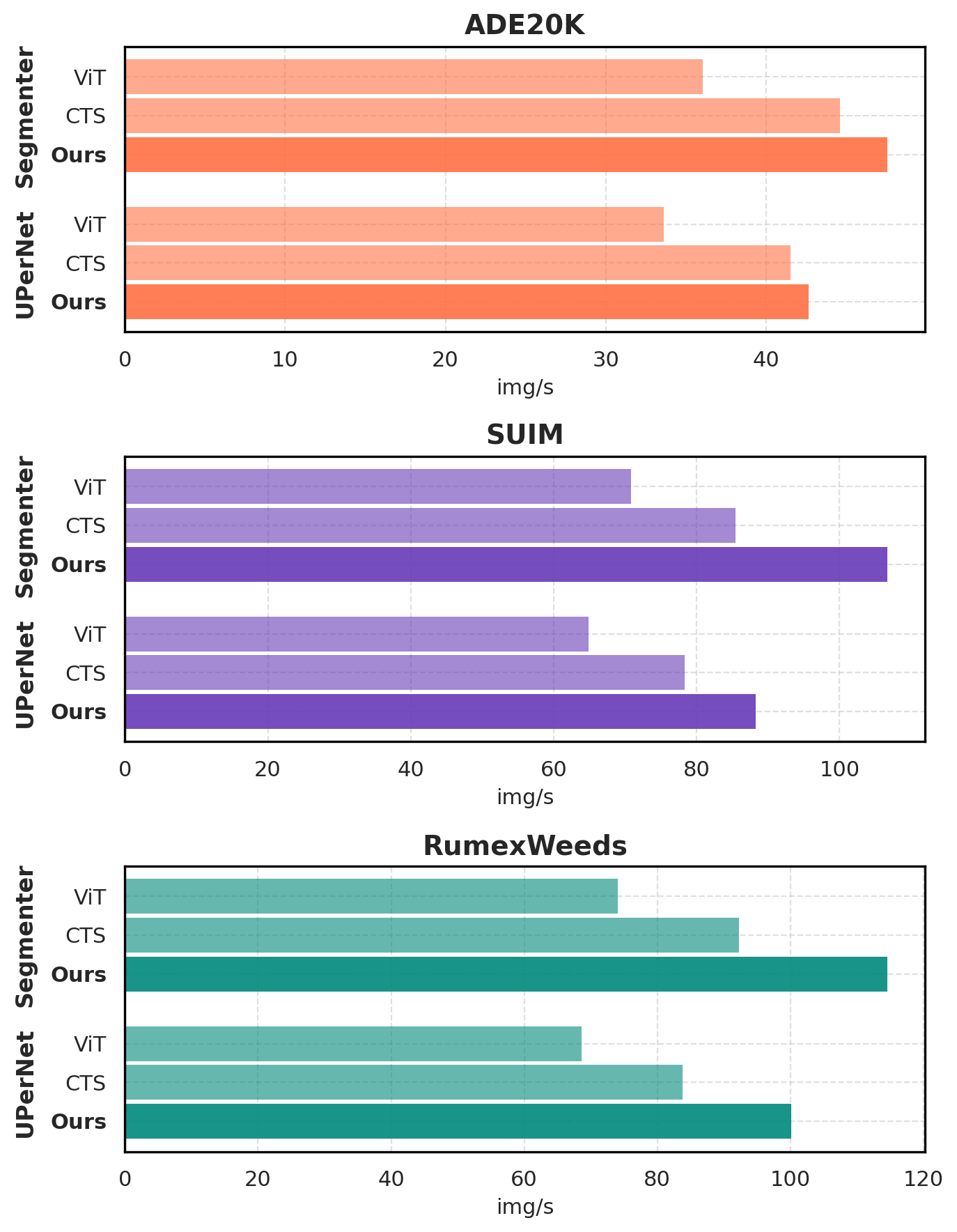}
  \caption{\textbf{Comparison of segmentation speed (\textit{img/s})} across three datasets (\textit{ADE20K}, \textit{SUIM}, and \textit{RumexWeeds}). Each plot shows results for different segmentation backbones: \textit{Segmenter} (top) and \textit{UPerNet} (bottom). For each dataset, we compare three models: \textit{ViT}, \textit{CTS}, and our model. Across both backbones and all datasets, our model consistently achieves the highest image throughput. The improvements are most pronounced for datasets with few subjects and dominated by background (see ablation study).}
  \label{fig:chart}
\end{figure}
\begin{figure*}[t]
    \centering
    \includegraphics[width=\textwidth]{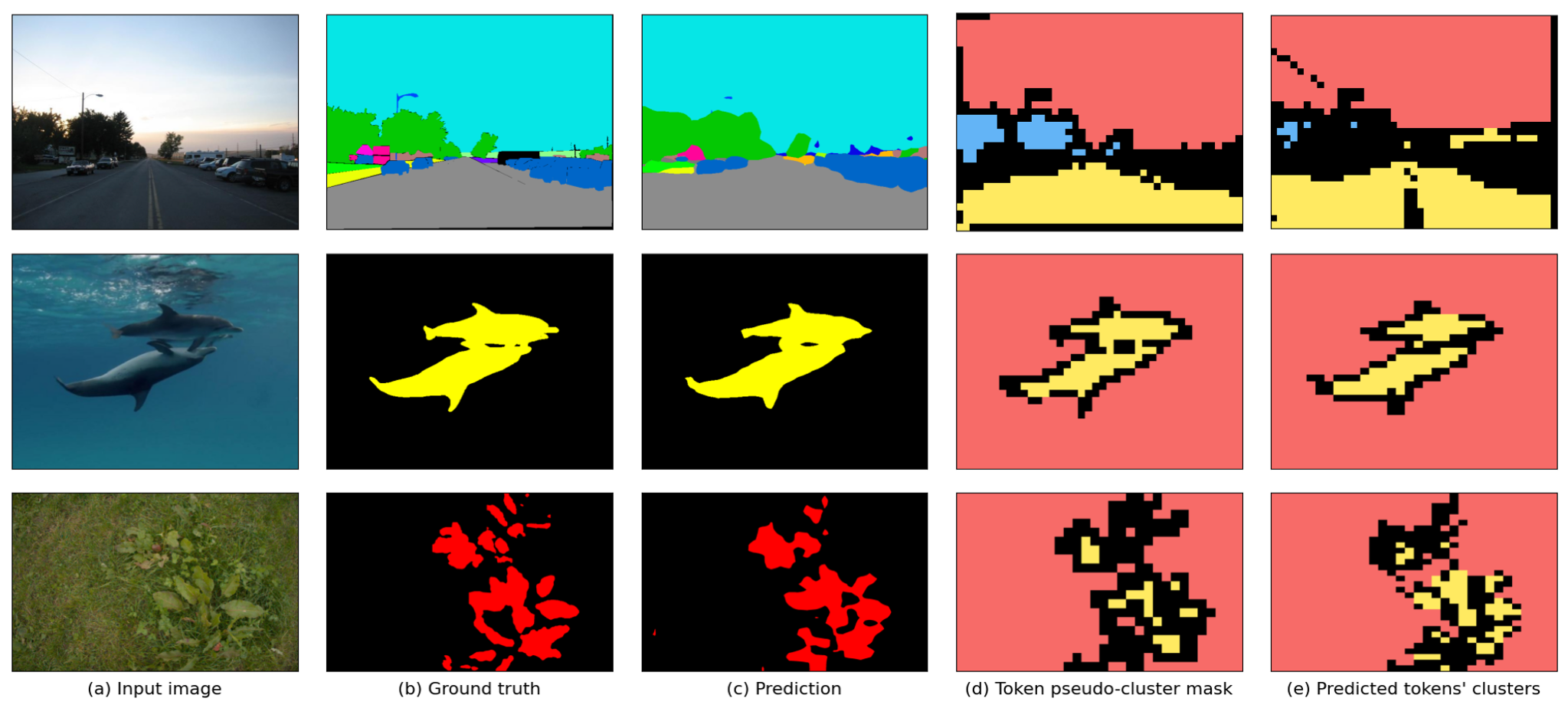} 
    \caption{\textbf{Examples from the \textit{ADE20K} \cite{zhouSceneParsingADE20K2017} (top), \textit{SUIM} \cite{islamSemanticSegmentationUnderwater2020} (middle), and \textit{RumexWeeds} \cite{RumexWeedsGrasslandDataset} (bottom) datasets}. Columns: (a) Input image, (b) Ground truth semantic segmentation, (c) Model prediction, (d) Mask for the token clustering generated from the ground truth, (e) Predicted cluster for each token. Starting from the output of (e), regions with the same non‑black color belong to the same cluster and get merged into a single token for the subsequent Transformer layers, while tokens in black regions are kept intact to preserve fine details. Our model configuration used to obtain these results is $\text{ClustViT-b}_{k3, ip3}$.}
    \label{fig:output_examples}
\end{figure*}
Our approach has several characteristics that make it well-suited for the task of reducing the computational complexity of ViTs in semantic segmentation. First, it introduces a clustering module that is trained end-to-end within the Transformer backbone, allowing the model to dynamically identify and merge semantically similar tokens during inference. Alike tokens get grouped from there on into representative tokens. This reduces the number of active tokens without disrupting the computation graph, leading to faster processing, as shown in Fig.~\ref{fig:chart}. 
Second, by leveraging pseudo-clusters derived from segmentation masks, the clustering process is guided by semantic information rather than low-level token similarity, which tailors the compression to the content of the image. Third, our architecture includes a regenerator module that reconstructs individual token representations from their clustered counterparts, ensuring compatibility with dense prediction heads. Together, these components enable efficient computation while maintaining high-quality segmentation outputs in a variety of datasets, as shown in Fig.~\ref{fig:output_examples}.
The \emph{contribution} of this work is threefold:
\begin{itemize}
    \item We present the concept of merging tokens according to the semantic content of the image---information that can be derived from the segmentation mask. This allows unstructured compression, simplifying processing.
    \item  We present ClustViT, an end-to-end trainable backbone that can identify and merge semantically similar tokens. We instill the standard ViT architecture with a component that clusters tokens based on image semantics. A regenerator component reconstructs tokens at the end of the encoder to enable the use of off-the-shelf segmentation heads. 
    \item Our approach significantly reduces the computational cost compared to state-of-the-art methods on datasets with few subjects and dominated by background---common in robotics---while achieving comparable segmentation accuracy on more complex datasets.
\end{itemize}

\section{Related Work}

\label{sec:related_works}
\subsection{Semantic Segmentation with Vision Transformers}
Semantic segmentation classifies each pixel in an image into its semantic category. It is a core problem in computer vision with applications in autonomous driving, image editing, robotics, and image analysis \cite{ronnebergerUNetConvolutionalNetworks2015, Gueldenring2021_IROS, jainOneFormerOneTransformer2023}. ViTs have recently emerged and significantly surpass previous convolutional approaches in various vision tasks, including semantic segmentation \cite{liTransformerBasedVisualSegmentation2024}. SETR \cite{zhengRethinkingSemanticSegmentation2021} is the first to successfully replace the traditional CNN backbone with a ViT backbone, though it still relies on a CNN decoder. Segmenter \cite{strudelSegmenterTransformerSemantic2021} showed that a pure encoder-decoder Transformer approach is a viable solution by designing a lightweight decoder mask Transformer that uses attention to produce class-specific masks. SegFormer \cite{xieSegFormerSimpleEfficient2021} improved scalability with a hierarchical Transformer encoder. SegViT \cite{zhangSegViTSemanticSegmentation} uses a plain ViT with the use of an attention mechanism for the creation of Transformer masks. Finally, Mask2Former \cite{chengMaskedattentionMaskTransformer2022a} introduces one architecture to address any possible segmentation task (panoptic, instance, or semantic) by constraining cross-attention to predicted mask regions.

\subsection{Token skimming background}
Token skimming starts from the assumption that not all parts of an image contribute equally to solving a task; many tokens contain irrelevant or redundant information. Therefore, skimming them can improve inference speed and help with signal noise reduction within the architecture. 
Token skimming can be done by either dropping tokens or \textit{merging} similar ones \cite{montelloSurveyDynamicNeural2025a}. 
This speedup approach has seen its biggest contributions in architectures for image classification.
DynamicViT \cite{raoDynamicViTEfficientVision2021a} prunes uninformative tokens progressively using a binary decision mask, applying hierarchical skimming across Transformer blocks. IA-RED$^2$ \cite{panIARED^2InterpretabilityAwareRedundancy2021} follows a similar approach but selects tokens via reinforcement learning \cite{williamsSimpleStatisticalGradientfollowing1992a}. SaiT \cite{liSaiTSparseVision2022a} dynamically drops tokens based on attention weights. GTP-ViT \cite{xuGTPViTEfficientVision2024a} performs token dropping as a per-block graph cut, optimizing towards the minimum normalized cut of tokens. Evo-ViT \cite{xuEvoViTSlowFastToken2022a} merges unnecessary tokens, summarizing the least important ones into placeholders. Finally, ToMe \cite{bolyaTokenMergingYour2023a} also combines similar tokens with a lightweight matching algorithm on the matrix $K$ of the attention block without requiring retraining.

\subsection{Token skimming for semantic segmentation}
Whereas most existing methods target image classification, some solutions have been proposed to address specifically segmentation. What makes dense prediction tasks different from classification is that all tokens are needed at the encoder output for it to be compatible with off-the-shelf segmentation heads.
Yuan et al. in \cite{yuanExpeditingLargeScaleVision2024a} present a token clustering layer and a token reconstruction layer into a pretrained ViT to merge and unmerge neighboring tokens based on a local $k$-means clustering. This method uses a structured per-block approach (fixed amount of compressed tokens). In our case, the focus is on tackling the problem based on semantic similarity instead of distance similarity among tokens.

The work of \cite{tangDynamicTokenPruning2023b} combines token skimming and early exits. After each Transformer block in the encoder, an auxiliary head halts \textit{simple} tokens, sending them to an early decoder, while the rest continue through the full encoder. This work is relevant to ours but approaches the problem differently by sending partial amounts of tokens earlier to the decoder head; each head sees only partial context at each exit. In contrast, we work on reconstructing all the tokens to be passed to the decoder head.

Most related to our approach, \cite{luContentawareTokenSharing2023b} proposes Content-aware Token Sharing (CTS) that uses a class-agnostic policy network attached before the tokenizer to predict whether $2 \times 2$ neighbor patches contain the same semantic class. In that case, the patch shares a token, effectively reducing resolution. The original resolution is then reconstructed after the backbone for CNN decoders or expanded after the decoder for Transformer decoders. 
This work is closest in spirit to our own approach, however, the token compression is structured (at most group of 4 tokens in each region) and predefined as a hyperparameter. In contrast, our method has the freedom of compressing an arbitrary amount of tokens. Furthermore, in CTS, a policy network needs to be trained aside from the main network. Our solution allows end-to-end training of the clustering component as a part of the architecture. 

Finally, different from all existing approaches, we explore clustering guided by prior knowledge from the segmentation mask, enabling token-level classification for clustering and merging similar ones. Our backbone is an optimized ViT, so comparisons naturally focus on this architecture.

\section{Method}
\begin{figure*}[t]
  \centering
  \includegraphics[width=\textwidth]{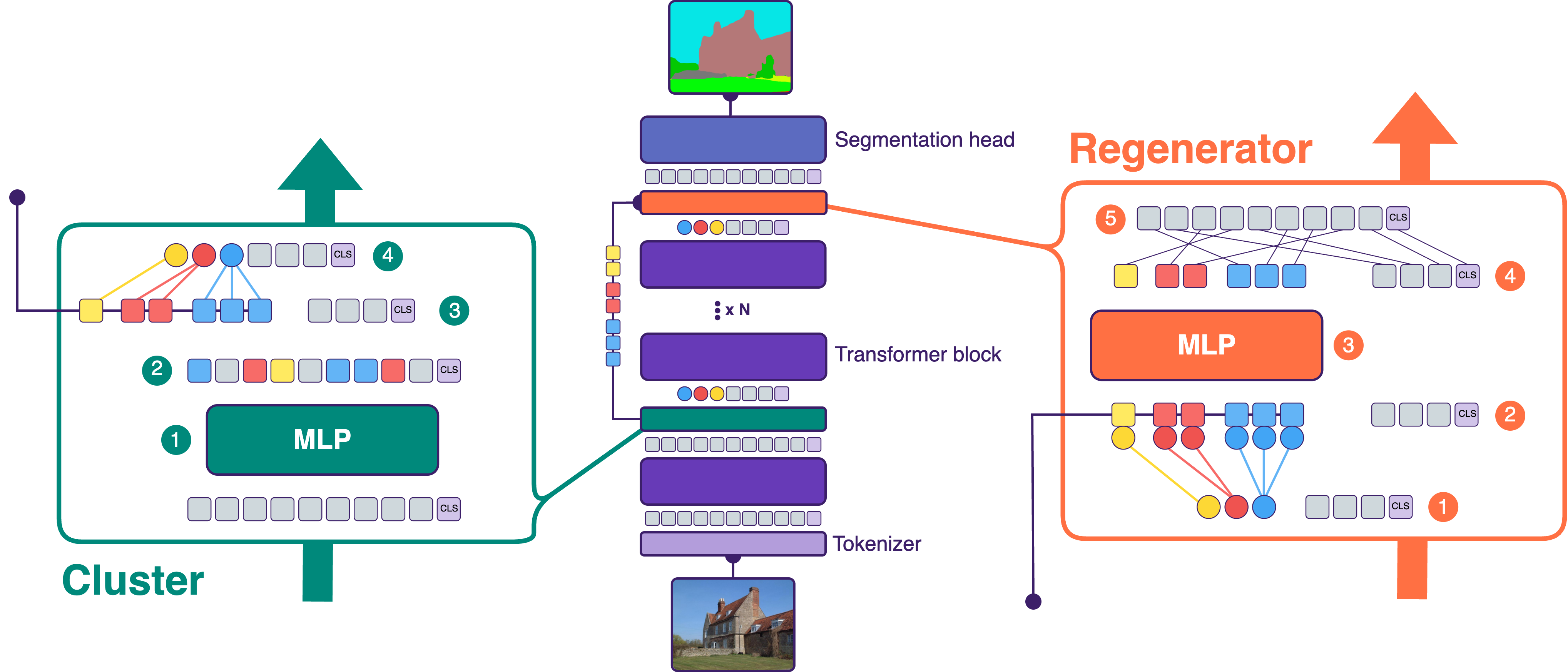}
  \caption{\textbf{ClustViT overview.} The standard Transformer pipeline is executed (center, from bottom to top) through the tokenizer and few Transformer blocks until the Cluster module is encountered. Subsequently, the Transformer backbone proceeds with a reduced amount of tokens. Before being passed to the segmentation head, the tokens are reconstructed by the Regenerator module. \textbf{Cluster module (left)}: \numberdotgreen{1} An MLP predicts the probability of a token belonging to a cluster. \numberdotgreen{2} Tokens of the same cluster (color coded) are grouped; unclustered (gray) tokens are kept intact. \numberdotgreen{3} Tokens within each group are aggregated into a single representative token. \numberdotgreen{4} The reduced token set (cluster representatives + unclustered tokens + CLS) is fed through the remaining Transformer blocks, lowering compute. \textbf{Regenerator module (right)}: \numberdotorange{1} Takes the reduced sequence. \numberdotorange{2} Uses stored assignments to expand each representative back to its original token positions. \numberdotorange{3} An MLP refines the reinstated per-token features. \numberdotorange{4} Reconstructed full-resolution tokens and preserved unclustered tokens are combined. \numberdotorange{5} The restored sequence is delivered to the segmentation head.}
  \label{fig:clustvit_architecture}
\end{figure*}
\label{sec:method}
In this section, we detail the core idea of clustering semantically similar tokens. We first briefly revisit the Transformer architecture (\ref{subsec:preliminaries}), followed by an overview of how our ClustViT method works overall (\ref{subsec:clustvit}). We then delve into details of the different ClustViT characteristics: the clustering module (\ref{subsec:clusterblock}), the regenerator module (\ref{subsec:regeneratorblock}), and the clustering training from the generation of pseudo-clusters and integration into a combined loss (\ref{subsec:clusteringloss}).

\subsection{Preliminaries}
\label{subsec:preliminaries}
Our approach expands upon the ViT \cite{dosovitskiyImageWorth16x162021b} architecture and leverages some of its unique properties. Assuming an RGB image $X \in \mathbb{R}^{H \times W \times 3}$ , ViT starts by splitting it into $P\times P$ non-overlapping patches, flattening them, and projecting each into a $D$-dimensional space. These projected patches, called tokens, are enriched with a positional encoding, and a specific classification token ($x_{cls}$) is concatenated at the beginning of the sequence to facilitate information flow. The resulting sequence is $Z \in \mathbb{R}^{(N + 1) \times D}$, where $N$ is the total number of patches. ViT then applies stacked attention blocks, each composed of Multi-Head Self-Attention (MHSA) and the Feed-Forward Network (FFN), both wrapped with Layer Normalization (LN) and residual connections. Subsequent attention blocks are called layers, indexed as $l = \{1, ..., N\}$, for which the complete set of operations in each block is the following:
\\
\begin{equation}
    \begin{aligned}
        Z'_l = \text{MHSA}(\text{LN}(Z_{l-1})) + Z_{l-1}
    \end{aligned}
\end{equation}
\begin{equation}
    \begin{aligned}
        Z_l = \text{FFN}(\text{LN}(Z'_l)) + Z'_l 
    \end{aligned}
\end{equation}
\\
ViT is commonly used as an encoder with task-specific decoders. For semantic segmentation, we experimented with two different segmentation heads: (\textit{i}) Segmenter \cite{strudelSegmenterTransformerSemantic2021}, which reprojects tokens into a latent space, concatenates learned class embeddings, and processes them with Transformer layers; and (\textit{ii}) UPerNet \cite{xiaoUnifiedPerceptualParsing2018}, which uses a feature pyramid and multi-scale fusion as its decoder, chosen for a fair comparison since it is part of the original encoder-decoder combination of CTS \cite{luContentawareTokenSharing2023b}. In both cases, outputs are upsampled by bilinear interpolation to the original resolution at the end.

\subsection{ClustViT}
\label{subsec:clustvit}
Our proposed ClustViT architecture modifies the standard ViT with a clustering mechanism to dynamically merge semantically similar tokens (cf. Fig.~\ref{fig:clustvit_architecture}). 
More precisely, in this work, we introduce two novel components to the standard ViT architecture: the Cluster and the Regenerator components. In between these two components, the Transformer blocks  remain unchanged but operate on a reduced (compressed) number of tokens.

As shown in Fig.~\ref{fig:clustvit_architecture}, the input image is split into patches, converted to tokens, augmented with positional embeddings, and prepended with the $x_{cls}$ token. The sequence is fed through the early layers of the encoder. All these steps are identical to the vanilla ViT. However, in our ClustViT, in between layers and at a predefined injection point ($ip_l$), the tokens enter our Cluster block (details in \ref{subsec:clusterblock}) which assigns each token to one of $k \in \mathbb{N}$ clusters or leaves it unclustered. Clustered tokens get aggregated into $k$ representative tokens, which attend subsequent standard Transformer blocks with the unclustered tokens, reducing sequence length and computational cost. After all encoder layers, our Regenerator block (details in \ref{subsec:regeneratorblock}) propagates updates from the representative tokens back to their original tokens. The restored sequence is then passed to the chosen off-the-shelf decoder head for prediction.

\subsection{Cluster Block}
\label{subsec:clusterblock}
This section introduces our cluster block (cf. left side of Fig.~\ref{fig:clustvit_architecture}), which compresses semantically similar tokens. 

MLP $\mathcal{C}$ (\numberdotgreen{1}) takes the tokens $Z_{l-1}$ as input and outputs cluster logits $L_C$. It is composed of two FFN with a ReLU activation in between:
\\
\begin{equation}
    \begin{aligned}
        L_C = \mathcal{C}(Z_{l-1}) = \text{Linear}_2(\text{ReLU}(\text{Linear}_1(Z_{l-1})))
    \end{aligned}
\end{equation}
\\
where $\text{Linear}_1: \mathbb{R}^{D} \to \mathbb{R}^{H}$ and $\text{Linear}_2: \mathbb{R}^{H} \to \mathbb{R}^{k+1}$, with $H$ being the hidden dimension and $k$ being a hyperparameter defining the number of clusters.\\
Thus, $L_C \in \mathbb{R}^{B \times N \times (k+1)}$. The output dimension $k+1$ corresponds to $k$ active clusters, plus one additional category for unclustered tokens.\\
The cluster assignments $C \in \{0, 1, \dots, k\}^{B \times N}$ are obtained by applying an $\operatorname{argmax}$ operation along the last dimension of the cluster logits $L_C$ (\numberdotgreen{2}). This assigns each patch token to a specific cluster ID (0 for unclustered, 1 to $k$ for active clusters):
\\
\begin{equation}
    \begin{aligned}
        C_{b,n} = \operatorname{argmax}_{k \in \{0, \dots, k\}} (L_C)_{b,n,k}
    \end{aligned}
\end{equation}
\\
where $b$ is the batch index and $n$ is the token index. \\
Based on the cluster assignments $C$, the tokens not belonging to any cluster are left untouched, while for each cluster $k$ the tokens get compacted into a representative token $E_k \in \mathbb{R}^{B \times D}$, computed as the mean of all patch tokens assigned to that cluster (\numberdotgreen{3} and \numberdotgreen{4}):
\\
\begin{equation}
    \begin{aligned}
        E_{b,k} &= \frac{1}{|\{n' \mid C_{b,n'} = k\}|} 
                  \sum_{n' \mid C_{b,n'} = k} Z_{l-1,b,n'} \\
        &\quad \text{where } |\{n' \mid C_{b,n'} = k\}| > 0
    \end{aligned}
\end{equation}
\\
The final reduced sequence to be passed to the subsequent Transformer layers is obtained by concatenating the kept patch tokens, and the representative tokens:
\\
\begin{equation}
    \begin{aligned}
        Z_{red} = [Z_{l-1,\text{unclustered}}; E]
    \end{aligned}
\end{equation}
\\
and it is of shape $Z_{red} \in \mathbb{R}^{B \times (N_{\text{unclustered}} + k) \times D}$. \\
The original matrix of the clustered tokens $Z_{l-1,\text{clustered}}$ is passed as residuals to the regenerator module for reconstruction with respect to the updated representatives. The classification token $x_{cls}$ is excluded during clustering and concatenated back afterward; this detail is omitted in the formalization and Fig.~\ref{fig:clustvit_architecture} for simplicity.

\subsection{Regenerator Block}
\label{subsec:regeneratorblock}
This section describes the steps to regenerate compressed tokens from the updated representatives and the residual clustered tokens (cf. right side of Fig.~\ref{fig:clustvit_architecture}).

Let $Z_{out}\in \mathbb{R}^{B \times (N_{\text{unclustered}} + k) \times D}$ be the output sequence from the Transformer layers after layer $l$ that processed the reduced sequence. We recover only the representative tokens by deconcatenating them from the unclustered tokens (\numberdotorange{1}). The two resulting matrices will be $Z_{repr}\in \mathbb{R}^{B \times k \times D}$ and $Z_{unclustered}\in \mathbb{R}^{B \times N_{\text{unclustered}} \times D}$. 

Subsequently, we expand $Z_{repr}$ in such a way that for each token in the residual matrix $Z_{l-1,\text{clustered}} \in \mathbb{R}^{B \times (N_{\text{clustered}} + k) \times D}$, the corresponding representative token from $Z_{repr}$ gets placed into an empty matrix $Z_{reprexp} \in \mathbb{R}^{B \times (N_{\text{clustered}} + k) \times D}$. We then proceed to concatenate the two matrices (\numberdotorange{2}) on the token embedding dimension $D$ and pass the resulting matrix to (\numberdotorange{3}) a refining MLP:
\\
\begin{equation}
    \begin{aligned}
        Z_{\text{refined}} = \text{Linear}_2(\text{GELU}(\text{Linear}_1(f_{\text{concat}})))
    \end{aligned}
\end{equation}
\\
where $\text{Linear}_1: \mathbb{R}^{2D} \to \mathbb{R}^{D}$ and $\text{Linear}_2: \mathbb{R}^{D} \to \mathbb{R}^{D}$. \\
Finally, we can recompose back the refined tokens into their original position before the merging. We take into account also the processed unclustered tokens (\numberdotorange{4}). Our final output sequence is $Z_{\text{final}} \in \mathbb{R}^{B \times (1+N) \times D}$ (\numberdotorange{5}). This sequence now matches the input sequence in length and is ready for the decoder head. As before, the classification token $x_{cls}$ is excluded during processing and concatenated back afterwards.

\begin{table}[t]
  \centering
  \caption{\textbf{Comparison of segmentation performance }across different heads (\textit{Segmenter}, \textit{UPerNet}) and backbones (\textit{ViT-b}, \textit{CTS-b}, \textit{ClustViT-b}) on \textit{ADE20K}, \textit{SUIM}, and \textit{RumexWeeds} datasets. Metrics reported include mIoU (higher is better), image throughput (higher is better), and GFLOPs (lower is better). Bold values indicate the best performance in each category. GFLOPs include standard deviation as ±.}
  \small
  \setlength{\tabcolsep}{2pt}
  \begin{tabular}{@{}llccc@{}}
    \toprule
    \textbf{Head} & 
    \textbf{Backbone} & 
    \shortstack{\textbf{mIoU ($\uparrow$)}} & 
    \shortstack{\textbf{img/s ($\uparrow$)}} & 
    \textbf{GFLOPs ($\downarrow$)} \\
    \midrule
    \multicolumn{5}{c}{\textbf{ADE20K}} \\
    \midrule
    \multirow{5}{*}{Segmenter} & ViT-b &  \textbf{49.22} & 36.06	& 473.15 ± 99.30 \\
     & CTS-b & 45.96 & 44.62 & 347.11 ± 72.85 \\
     & $\text{ClustViT-b}_{k3, ip3}$ & 46.10 & \textbf{47.56} & \textbf{321.28 ± 88.83} \\
     & $\text{ClustViT-b}_{k3, ip4}$ & 46.19 & 45.52 & 338.68 ± 89.91 \\
     & $\text{ClustViT-b}_{k1, ip4}$ & 48.20 & 41.66 & 399.13	± 99.86 \\
    \cmidrule(lr){2-5}
    \multirow{5}{*}{UPerNet} & ViT-b & \textbf{47.53} & 33.63 & 637.51 ± 72.85 \\
    & CTS-b & 46.85 & 41.55 & 511.48 ± 107.35 \\
    & $\text{ClustViT-b}_{k3, ip3}$ & 44.70 & \textbf{42.65} & \textbf{494.67 ± 119.59}  \\
    & $\text{ClustViT-b}_{k3, ip4}$ & 44.16 & 41.64 & 508.52 ± 120.07 \\
    & $\text{ClustViT-b}_{k1, ip4}$ & 45.92 & 37.69 & 566.97 ± 131.10 \\
    \midrule
    \multicolumn{5}{c}{\textbf{SUIM}} \\
    \midrule
    \multirow{5}{*}{Segmenter} & ViT-b & \textbf{69.91} & 70.80 & 252.33 ± 0.00 \\
     & CTS-b & 66.33 & 85.41 & 183.40 ± 0.00 \\
     & $\text{ClustViT-b}_{k3, ip3}$ & 61.95 & 106.65 & \textbf{115.49 ± 9.00} \\
     & $\text{ClustViT-b}_{k4, ip3}$ & 63.26 & 115.95 &134.19 ± 	14.54 \\
     & $\text{ClustViT-b}_{k4, ip4}$ & 63.86 & \textbf{116.56} & 132.94 ± 8.95 \\
    \cmidrule(lr){2-5}
    \multirow{5}{*}{UPerNet} & ViT-b & \textbf{70.01} & 64.9 & 384.24 ± 0.00 \\
    & CTS-b & 68.05 & 78.31 & 279.35 ± 0.00 \\
    & $\text{ClustViT-b}_{k3, ip3}$ & 63.77 & 88.31 & \textbf{221.02 ± 15.04} \\
    & $\text{ClustViT-b}_{k3, ip4}$ & 64.98 & 97.86 & 236.53 ± 8.30\\
    & $\text{ClustViT-b}_{k2, ip4}$ & 65.02 & \textbf{103.54} & 224.51 ± 16.06\\
    \midrule
    \multicolumn{5}{c}
    {\textbf{RumexWeeds}} \\
    \midrule
    \multirow{5}{*}{Segmenter} & ViT-b & 51.29 & 74.10 & 252.07 ± 0.00 \\
     & CTS-b & 48.54 & 92.32 & 183.18 ± 0.00\\
     & $\text{ClustViT-b}_{k3, ip3}$  & 49.82 & \textbf{114.53} & 122.23 ± 46.76 \\
     & $\text{ClustViT-b}_{k2, ip3}$  & 50.45 & 115.62 & \textbf{118.73 ± 41.80} \\
     & $\text{ClustViT-b}_{k3, ip4}$  & \textbf{51.53} & 107.34 & 135.54 ± 	37.98 \\
    \cmidrule(lr){2-5}
    \multirow{5}{*}{UPerNet} & ViT-b & \textbf{51.56} & 68.61 & 348.23 ± 0.00 \\
    & CTS-b & 49.13 & 83.76 & 279.34 ± 0.00 \\
    & $\text{ClustViT-b}_{k3, ip3}$  & 50.55 & \textbf{100.18} & 221.14 ± 34.20 \\
    & $\text{ClustViT-b}_{k2, ip3}$  & 49.63 & 99.99 & \textbf{221.00 ± 	34.22} \\
    & $\text{ClustViT-b}_{k3, ip4}$  & 51.20 & 94.17 & 237.91 ± 	35.87 \\
    
    \bottomrule
  \end{tabular}
  \label{tab:datasets_results}
\end{table}

\subsection{Pseudo-clusters and combined loss for model training}
\label{subsec:clusteringloss}
To train the cluster module MLP, we generate pseudo-clusters from the segmentation mask. We split the segmentation mask into patches, as done by the tokenizer. For each patch, we check if all the pixels inside belong to the same semantic class. If all pixels belong to the same semantic class, that class is assigned; otherwise, a value of $0$ indicates mixed classes. Next, among patches of a single class, we keep only the $\text{top-}k$ most frequent classes, where $k$ is the hyperparameter for the number of clusters. Classes outside of the $\text{top-}k$ are set to $0$; $\text{top-}k$ classes get a label number $l \in {1,...,k}$, ordered from the most to the least frequent. Examples of pseudo-clusters are shown in Fig.~\ref{fig:output_examples}(d).
\par Once pseudo-clusters are computed, we define a composite loss that accounts for both segmentation accuracy and cluster behavior. For both tasks, we use cross-entropy: 
\\
\begin{equation}
    \begin{aligned}
        \mathcal{L} = -\frac{1}{N} \sum_{n=1}^{N} \sum_{c=1}^{C} y_{n,c} \log(\hat{y}_{n,c})
    \end{aligned}
\end{equation}
\\
where for the segmentation case $N$ is the number of pixels, $C$ is the number of classes, $y_{n,c}$ is the ground truth, while $\hat{y}_{n,c}$ is the predicted class value. In the case of clustering, $N$ is the total number of patches, $C = k+1$, where $k$ is the number of clusters; $y_{n,c}$ is the pseudo-cluster for a specific patch, and $\hat{y}_{n,c}$ is the predicted cluster.
The final loss is then composed of:
\\
\begin{equation}
    \begin{aligned}
        \mathcal{L} = \mathcal{L}_{segm} + \lambda\,\mathcal{L}_{clust}
    \end{aligned}
\end{equation}
\\
where $\lambda$ balances the relative importance of the clustering loss.

\section{Experiments}
\begin{table}[t]
  \centering
  \caption{\textbf{Ablation study on the impact of cluster count ($k$) and injection point ($ip$)} in \textit{ClustViT-b} evaluated on \textit{ADE20K} and \textit{Segmenter} head. Metrics include mIoU (higher is better), image throughput (higher is better), and GFLOPs (lower is better). \textit{ViT-b} serves as the non-clustered baseline. GFLOPs include standard deviation as ±.}
  \small
  \setlength{\tabcolsep}{6pt}
  \begin{tabular}{@{}lccrrl@{}}
    \toprule
    \textbf{Backbone} & \textbf{\textit{k}} & \textbf{\textit{ip}} &
    \textbf{mIoU ($\uparrow$}) & \textbf{img/s ($\uparrow$)} & \textbf{GFLOPs ($\downarrow$)} \\
    \midrule
    ViT-b & -- & -- & 49.22 & 36.06 & 473.15 ± 99.30 \\
    \midrule
    \multirow{5}{*}{ClustViT-b} & 1 & \multirow{5}{*}{4} & \textbf{48.20} & 41.66 & 399.13 ± 99.86 \\
                                & 2 &                         & 47.32 & 44.59 & 356.09 ± 94.79 \\
                                & 3 &                         & 46.19 & 45.52 & 338.68 ± 89.91 \\
                                & 4 &                         & 46.13 & 46.73 & 326.53 ± 85.31 \\
                                & 5 &                         & 44.82 & \textbf{46.87} & \textbf{321.30 ± 82.60} \\
    \midrule
    \multirow{5}{*}{ClustViT-b} & \multirow{5}{*}{3} & 2 & 45.00 & \textbf{50.23} & \textbf{304.49 ± 87.16} \\
                                &                     & 3 & 46.49 & 48.07 & 319.78 ± 88.90 \\
                                &                     & 4 & 46.19 & 45.52 & 338.68 ± 89.91 \\
                                &                     & 5 & \textbf{46.70} & 44.13 & 354.42 ± 88.30 \\
                                &                     & 6 & 46.25 & 42.29 & 370.80 ± 88.53 \\
    \bottomrule
  \end{tabular}
   \label{tab:ablation_results}
\end{table}

\subsection{Datasets and Metrics}
We evaluate our backbone on three different semantic segmentation datasets. The first one, \textbf{ADE20K} \cite{zhouSceneParsingADE20K2017}, contains \textit{20k} training and \textit{2k} validation images across 150 classes in indoor and outdoor scenes; results are reported on the validation set. While \textit{ADE20K} tests generalizability, it is less representative of real-world robotics scenarios. For this reason, we also use the \textbf{SUIM} dataset \cite{islamSemanticSegmentationUnderwater2020} containing underwater scenes and \textbf{RumexWeeds} \cite{RumexWeedsGrasslandDataset} containing agricultural images.
SUIM is a dataset for underwater robotics composed of \textit{1525} training images (split 85/15 for test and validation) and \textit{110} test images, with 8 different classes.
RumexWeeds is composed of data collected in grasslands and targets two weed species with \textit{2796} training, \textit{1411} validation and \textit{1303} test images.
\par Segmentation accuracy is measured via mean Intersection over Union (\textbf{mIoU}). Inference speed is reported as image throughput (\textbf{img/s}) and Giga Floating Point Operations (\textbf{GFLOPs}) are calculated to estimate the computational cost; in the case of GFLOPs, standard deviation is included to reflect fluctuations from model dynamics and input image size. It remains fixed with respect to repeated iterations (in contrast to \textit{img/s} that requires warmup).
\begin{table}[t]
  \centering
  \caption{\textbf{Performance comparison of \textit{Segmenter} head at varying model scales (tiny, small, large)}. Reported metrics include mIoU (higher is better), image throughput (higher is better), and GFLOPs (lower is better). GFLOPs include standard deviation as ±.}
  \small
  \begin{tabular}{@{}lccc@{}}
    \toprule 
    \textbf{Backbone} & 
    \shortstack{\textbf{mIoU ($\uparrow$)}} & 
    \shortstack{\textbf{img/s ($\uparrow$)}} & 
    \textbf{GFLOPs ($\downarrow$)} \\
    \midrule
     ViT-t & \textbf{38.62} & \textbf{96.55} & \textbf{46.27 ± 9.71} \\
     $\text{ClustViT-t}_{k3, ip4}$ & 36.40 & 91.39 & 31.13 ± 8.13\\
    \midrule
     ViT-s &  \textbf{45.58} & 66.66 & 140.55 ± 29.50\\
     $\text{ClustViT-s}_{k3, ip4}$ &  42.76 & \textbf{75.01} & \textbf{96.26 ± 25.19}\\
    \midrule
      ViT-l &  \textbf{51.45} & 8.48 & 2455.02 ± 514.21 \\
     $\text{ClustViT-l}_{k3, ip4}$ &  49.93 & \textbf{15.58} & \textbf{1363.08 ± 444.4} \\
    \bottomrule
  \end{tabular}
  \label{tab:segmenter_vit}
\end{table}
\subsection{Experimental setup}
\par \textbf{Model and loss configuration.} We use the base ViT architecture for model comparisons and hyperparameter ablation, while also showing how our optimized solution scales across different ViT sizes. All architectures use a patch size of 16. Our architecture adds a clustering MLP to the original ViT parameters, with hidden dimensions of 774 (tiny), 1548 (small), 3096 (base), and 4128 (large). For the combined loss, after empirical testing, we found that $\lambda=0.1$ works best to balance the resolution of both segmentation and clustering problems.
As mentioned, we compare directly with CTS \cite{luContentawareTokenSharing2023b}, considered the state-of-the-art architecture. While our approach could support full-network token compression (including both encoder and decoder), as seen in CTS, we focus only on the backbone to allow standard segmentation heads to be used interchangeably. Among CTS’s token-sharing settings, we adopt the configuration achieving the highest accuracy, which compresses 30\% of tokens.
\par \textbf{Training setup.} We use \textit{mmsegmentation}\footnote{\hyperlink{https://github.com/open-mmlab/mmsegmentation}{https://github.com/open-mmlab/mmsegmentation}} as base framework. Following Segmenter \cite{strudelSegmenterTransformerSemantic2021} we train with SGD (learning rate 0.001, momentum 0.9, weight decay 0.0005) and a polynomial decay of power 0.9 to a minimum of 0.0001 over 160k iterations for \textit{ADE20K} and 80k for \textit{RumexWeeds} and \textit{SUIM}. Batch size is 8 during training; inference is on single images. All \textit{mIoU} scores use single-scale inputs. \textit{ADE20K} images retain their original resolution, while \textit{RumexWeeds} and \textit{SUIM} images are resized to $640\times640$. Backbones are initialized with ImageNet-pretrained \cite{dengImageNetLargescaleHierarchical2009} ViT weights ($384\times384$, patch size 16) from \textit{mmsegmentation}. Our approach supports multiple segmentation heads: Segmenter (transformer-based) and UPerNet (CNN-based). Segmenter uses only the ViT output tokens, which we provide after reconstruction. UPerNet also uses intermediate tokens (after layers 4, 6, and 8 for base ViT), which are reconstructed at each stage via the regenerator module.

\par \textbf{Infrastructure.}{ All trainings are performed on a server on a single Nvidia H100 GPU with 80GB of VRAM, while image throughput and GFLOPs were measured on a workstation with an Intel i9-14900KF CPU, 96GB of RAM and a Nvidia 3090 GPU with 24GB of VRAM.}

\begin{figure}[t]
  \centering
  \includegraphics[width=\linewidth]{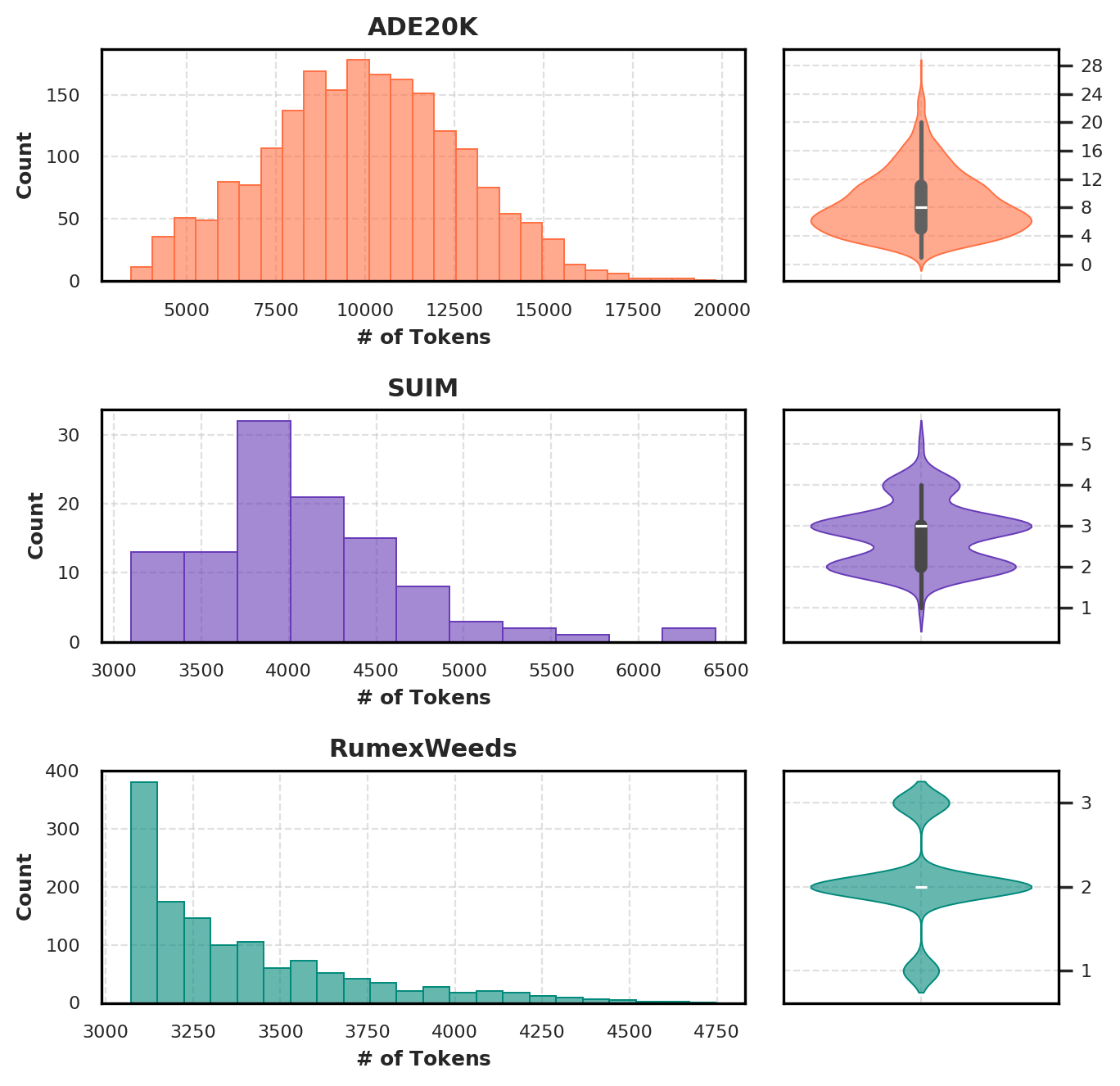}
  \caption{\textbf{Distribution of token counts and class diversity across test sets}. Each row shows the histogram of tokens used by $\text{ClustViT-b}_{k3, ip3}$ (left) and the average number of classes per image (right). \textit{ADE20K} exhibits a symmetric token distribution being a dataset with high class diversity, \textit{SUIM} is moderately left-skewed being of moderate diversity, while \textit{RumexWeeds} is sharply peaked and is composed of low class diversity images.}
  \label{fig:token_dist}
\end{figure}

\subsection{Results}
Table \ref{tab:datasets_results} reports results for different backbones and heads across the three datasets, showing the top three configurations per dataset in terms of \textit{mIoU}. We provide different configurations of our ClusterViT architecture based on the number of clusters $k$ and the injection point $ip$ where tokens are compressed. For example, ClustViT-b$_{k4,ip2}$ is trained to perform at most 4 clusters ($k=4$) and the Cluster module is placed after the second Transformer block ($ip=2$) in the architecture.  On \textit{ADE20K}, ClustViT-b$_{k3,ip3}$ achieves the lowest computational cost ($321.28$ GFLOPs), a $1.47\times$ improvement over the baseline, with a trade-off in terms of accuracy of $1.66$ mIoU points (about $3.7\%$). The best-performing model still achieves $1.15\times$ speedup with respect to the baseline, with only $2\%$ accuracy decrease. Using UPerNet as the head shows a similar pattern, although the accuracy seems to be affected more by our compression method.

On \textit{SUIM}, which contains large background areas (water), Segmenter predictions decrease notably: ClustViT-b$_{k4,ip3}$ drops in \textit{mIoU} of $9.5\%$ compared to the baseline and $4.6\%$ compared to CTS, likely due to token merging reducing details representation. However, speed improvements are substantial: $1.64\times$ speedup in terms of FPS and $2.18\times$ increase in terms of GFLOPs in the respective best cases. Even when compared to CTS, we can observe $+21$ FPS increase and $1.58\times$ increase of GFLOPs. UPerNet confirms this trend. Notably, fastest FPS does not always correspond to lowest GFLOPs, likely due to architectural optimizations for specific operations, which can run in parallel and be aggregated.

On \textit{RumexWeeds}, a dataset composed mostly of grassy background, Segmenter slightly outperforms even the baseline ($+0.24\%$ in mIoU) while achieving $1.54\times$ FPS and $2.12\times$ GFLOPs gains. This highlights ClustViT’s strength in scenarios with few subjects and largely uniform backgrounds, common in many robotics applications.

To further understand the accuracy trade-offs of token clustering, we evaluated global size-stratified recall (Table~\ref{tab:size_stratified_results}) at different sizes relative to the image area:  Small ($<$5\%), Medium (5--15\%), Large (15--30\%), and Huge ($>$30\%). While \textit{ClustViT-b} maintains near-baseline accuracy on ``Huge'' elements, it causes a minor consistent performance drop across ``Small,'' ``Medium,'' and ``Large'' objects, instead of disproportionately penalizing smaller classes.

\subsection{Ablations}
\begin{table}[t]
  \centering
  \caption{\textbf{Global size-stratified recall evaluation} between \textit{ViT-b} and $\text{ClustViT-b}_{k3, ip3}$ across the datasets. Values represent the recall percentage for each relative object size category. Missing ground truth is indicated with "--".}
  \small
  \setlength{\tabcolsep}{4pt}
  \begin{tabular}{@{}lcccc@{}}
    \toprule
    \multirow{2}{*}{\textbf{Model}} & \textbf{Small} & \textbf{Medium} & \textbf{Large} & \textbf{Huge} \\
    & \scriptsize($<$5\%) & \scriptsize(5--15\%) & \scriptsize(15--30\%) & \scriptsize($>$30\%) \\
    \midrule
    \multicolumn{5}{c}{\textit{ADE20K}} \\
    \midrule
    ViT-b          & 65.99 & 81.06 & 86.45 & 88.57 \\
    ClustViT-b-3,3 & 63.48 & 78.93 & 84.84 & 87.52 \\
    \midrule
    \multicolumn{5}{c}{\textit{SUIM}} \\
    \midrule
    ViT-b          & 68.25 & 82.50 & 87.01 & 92.70 \\
    ClustViT-b-3,3 & 63.60 & 73.67 & 77.68 & 91.42 \\
    \midrule
    \multicolumn{5}{c}{\textit{RumexWeeds}} \\
    \midrule
    ViT-b          & 49.21 & 87.26 & -- & 99.51 \\
    ClustViT-b-3,3 & 45.09 & 84.25 & -- & 99.50 \\
    \bottomrule
  \end{tabular}
   \label{tab:size_stratified_results}
\end{table}

Table \ref{tab:ablation_results} reports the ablation results versus the original ViT backbone, showing how accuracy and performance vary with the number of clusters $k$ and the injection point $ip$ .
\par\textbf{Cluster size.} Fixing $ip=4$ and varying $k$ from 1 to 5, we see that increasing clusters reduces \textit{mIoU} while improving efficiency. For $k=1$, \textit{mIoU} is $48.20$ (close to the ViT-b baseline of $49.22$), with $15.5\%$ higher throughput and $1.19\times$ lower GFLOPs. Increasing to $k=5$ raises FPS by $29.9\%$ but drops \textit{mIoU} to $44.82$. This trade-off arises because larger clusters preserve fewer details: small objects are clustered and lose fine-grained information, while a single cluster leaves most tokens unclustered.
\par \textbf{Clustering positioning.} Fixing $k=3$ and varying $ip$ from 2 to 6 shows that earlier compression increases efficiency but can reduce accuracy. For instance, $ip=2$ yields the highest throughput ($50.23$ img/s) and lowest GFLOPs ($304.49$) but the lowest \textit{mIoU} (45.00). Interestingly, $k=3,ip=3$ outperforms $ip=4$ in both speed and \textit{mIoU}, suggesting that compressing tokens earlier benefits computational efficiency, while certain embedding spaces are more adapted to perform clustering with higher quality.
\par \textbf{Model size.} Table \ref{tab:segmenter_vit} shows the performance of the model for different sizes of the backbone. The added operations to perform the clustering penalize the smallest version, where the trade-off with the removed tokens is not favorable. The larger the model gets, the larger the performance gains are, and the closer the \textit{mIoU} is to the baseline.
\par \textbf{Tokens distribution.} Figure~\ref{fig:token_dist} shows the distribution of the number of tokens processed for each image by $\text{ClustViT-b}_{k3, ip3}$ at inference time on the respective test sets, alongside the distribution of the average number of classes per image. The \textit{ADE20K} histogram has the widest spread and is very symmetric to the center, reflecting high token variability and differing image resolutions (kept unscaled during inference). \textit{SUIM} is more left-skewed, with most images using fewer tokens than average and a few requiring many more. \textit{RumexWeeds}, dominated by background, shows the least variation: most images use very few tokens, with only a handful needing above-average counts to handle local complexity. A correlation between class diversity and token distribution can be observed. Speedups are most pronounced on datasets like \textit{RumexWeeds} where scenes with fewer classes lead to simpler segmentations, allowing for stronger compression in architectures where the amount of tokens that can be compressed is unbounded. In contrast, high-diversity datasets, like \textit{ADE20K}, require more tokens, resulting in speedup in line with other state-of-the-art solutions.

\section{Conclusions}

In this paper, we presented a novel approach to token compression that exploits semantic information from segmentation masks to guide the token merging process. Building on top of the ViT backbone, we introduced ClustViT, a model that integrates a clustering mechanism to adaptively merge tokens in between Transformer layers. A regenerator block is also introduced to restore the full representation for compatibility with standard segmentation heads. Through our experiments, we showed that this method yields significant computational savings on datasets with few objects and large background areas---conditions that are common in many robotic applications---while still maintaining competitive segmentation accuracy on visually complex datasets. These results highlight how semantically guided token compression improves the tradeoff between efficiency and accuracy, and point toward broader opportunities for incorporating token clustering strategies into vision models for robotics.

\bibliographystyle{IEEEtran}
\bibliography{IEEEabrv, main}

\end{document}